\documentclass[conference]{IEEEtran}

\usepackage{url}
\usepackage{blindtext, graphicx}
\usepackage{cite}

\ifCLASSINFOpdf
\else
\fi
\title{SROS: Securing ROS over the wire, \\
in the graph, and through the kernel}

\author{
    \IEEEauthorblockN{Ruffin White and Dr. Henrik I. Christensen}
        \IEEEauthorblockA{Contextual Robotics Institute\\
        UC San Diego, California, USA} 
    \and
    \IEEEauthorblockN{Dr. Morgan Quigley}
        \IEEEauthorblockA{Open Source Robotics Foundation\\
        Mountain View, California, USA} 
}

\begin{document}

\maketitle

\begin{abstract}
SROS is a proposed addition to the ROS API and ecosystem to support modern cryptography and security measures. An overview of current progress will be presented, rationalizing each major advancement, including: over-the-wire cryptography for all data transport, namespaced access control enforcing graph policies/restrictions, and finally process profiles using Linux Security Modules to harden a node’s resource access. By making the community aware of the vulnerabilities in ROS, as well as the proposed solutions provided by SROS, we intend to improve the state of security for future robotics subsystems.

\end{abstract}


\section{Introduction}

Cybersecurity is quickly becoming a pervasive issue for robotics, especially so as robots become more ubiquitous within society. With the advent of industrial automation, autonomous vehicles, robot-assisted surgery, commercial surveillance platforms, home service robots, and many more robotics domains, security of these subsystems should be considered vital, as they all provide a vector for cyber threats to manifest into real-world risks. Even without the hazards associated with industrial-strength robot arms or high-speed driverless semi trucks, personal robots promising to integrate with the Internet Of Things could become targets for breaches in privacy and sources of identity theft, similar to smartphones and PCs \protect\cite{lera2016cybersecurity}.

The Robot Operating System (ROS), a widely adopted standard robotic middle-ware due in part to its active community, provides a communication layer abstracted above a host operating system to construct a heterogeneous compute cluster for robots \protect\cite{quigley2009ros}. ROS, an open source initiative, was developed to simplify code reuse among robots with wildly varying hardware and to support large-scale software integration efforts.

However, original development of ROS was biased by attributes valued by robotics researchers, including: flexible computational graphs, modular publish and subscribe networks, and rapid software prototyping. Now that ROS and deriving robot platforms are out-growing the realm of research and into commercial \& industrial sectors, support for network security, identity authorization, and scoping resource permissions have quickly risen to the forefront of requested features.

To answer these requests, we propose a set of new security features to the core of ROS's codebase, Secure ROS (SROS). Announced publicly at ROSCon 2016 by authors White and Quigley\protect\cite{white16roscon}, SROS is a proposed addition to the ROS API and ecosystem to support modern cryptography and security measures in an effort to address existing vulnerabilities.

Current features in SROS include: native TLS support for all IP/socket level communication within ROS, the use of x.509 certificates permitting chains of trust, definable namespace globbing for ROS node restrictions and permitted roles, as well as convenient user-space tooling to auto generate node key pairs, audit ROS networks, and construct/train access control policies. In addition, AppArmor profile templates will also be introduced, allowing users to harden or quarantine ROS based processes running on Linux.

\section{Secure Communications}

In ROS, nodes intercommunicate through an API via XML-RPC, a remote procedure call protocol using XML encoding, as well as message/service data exchanges using transport libraries such as ROSTCP or ROSUDP for serialization over IP sockets. A glaring deficiency in the traditional infrastructure is the fact that all network traffic is transmitted in clear text. Additionally, no integrity checking is preformed on received packets other than basic message type continuity and API call validity, i.e. there are no means to verify payload information was unaltered in transit. This makes ROS a prime target for packet sniffing and man-in-the-middle attacks, resulting in an absence of native network confidentiality and data integrity.

In SROS, all network communication is encrypted using Secure Sockets Layer (SSL), or more specifically Transport Layer Security (TLS). This done through the use of Public Key Infrastructure (PKI), where by each ROS node is provided an X.509 certificate, equivalently an asymmetric key pair, signed by a trusted certificate authority (CA). These additions are all done at the ROS library level, enabling users with preexisting code and projects levering ROS's abstraction layers to also seamlessly support SROS.

These additional safety measures do bring their own complexities, but as with the authors' opinion, security without usability remains insecure, we've also developed tooling to support and simplify the tasks of using PKI in ROS. SROS provides a keyserver for certificate generation, as well as native API to distribute cyphered certificates to nodes upon a setup. The keyserver remains separate from roscore, and can thus easily be executed elsewhere on the network or taken offline completely, elevating the need for CA exist on the robot, leaving certificates imitable. Additionally, the keyserver provides a customizable configuration, where users can curtail certificate and CA properties by a node or node name-space basses, e.g. key algorithm, bit length, fingerprint, as well as CA info, extensions, restrictions, hierarchy, etc.

\section{Access Control}

ROS uses name-spaces to define most locations and aspects of ROS's computational graph, such as message topics, services, parameters, and the resolvable names of nodes themselves. Besides the requirement that nodes be named uniquely, the traditional infrastructure provides no level of access control for graph actions such as limiting what name-spaces a node may assume, what topics it can publish/subscribe, parameters it may read/write, services it may call/advertise, or what internal ROS API it may invoke in the graph. Although this flexible aspect of ROS makes it pleasant for debugging and rapid prototyping, the absence of any safeguards leaves ROS graphs susceptible to rogue or compromised nodes, spoofing message data and master/slave functions such as name-space registration or node shutdown. This also makes the use of ROS in industry tricky, as there are no guaranties of maintaining graph topology once deployed.

SROS introduces access control through the use of PKI by embedding policy definitions in X.509 certificate extensions. Again, because these certificates are signed, any attempt to mutate or elevate a node's own permissions will void the CA's signature, and thus fail during the TLS handshake for incoming and outgoing p2p connections. A node's policy is defined by the presence of object identifier (OID) fields, where each OID maps to specific allowed or denied action, and the value of the field uses a regex like syntax to define the name-space scope of the action in question. This policy profiling resembles that of AppArmor's mandatory access control (MAC) \protect\cite{bauer2006paranoid}, where permissions may be aliased across a sub-tree path, but then also revoked for specific nested sub-trees. In fact, SROS's default globbing style is the same used in AppArmor, as the syntax is more applicable to scoping paths and is human readable for auditing purposes.

This security feature also introduces its own complexities, since flawlessly defining a policy profile that encapsulates all of the expected exchanges in a ROS graph would be a daunting task manually. So in line with the authors' focus on security and usability, tooling for learning policy profile through demonstration has also been incorporated into SROS. SROS provides a varying degree of run-time modes, including audit, enforce and complain, again borrowing feature designs from AppArmor. This provides developers a method to auto generate, or amend profiles through granular logging of access events and violation attempts.

\section{Process Profiles}

For networked systems, application security is arguably the single-most important area of concern \protect\cite{bauer2006paranoid}, and as ROS sits squarely in this domain as a networking middle-ware for robotic applications, all facets of its functions should be secured. However ROS packages come from varying sources, and vetting each dependency in this high-level stack for uncommon vulnerabilities can be impractical for many. Enabling MAC for ROS nodes processes can help mitigate zero-day-exploits that enable attackers to gain shell access or invoke local privilege escalation, as well as quarantine unknown malfunctions during run-time such as accidentally overwriting another node's log files.

AppArmor is one such implementation of MAC building from the security modules available in the Linux kernel, and is widely adopted in Ubuntu, ROS's primary release target. AppArmor is also well documented and user-friendly relative to alternatives such as SELinux, making it a suitable MAC of choice for ROS. SROS provides a profile library for AppArmor, composed of modular primitives to quickly build custom profiles for ROS nodes. These include the minimal permissions necessary for core ROS features, such as the interposes signalling needed for roslaunch to manage nodes, shared library access for nodes written in python or C++, network access for socket communication, etc. Using the SROS profile library, users can focus on defining MAC pertinent to the application, e.g. prescribing nodes access to a serial buss, camera peripherals, or e-stop interfaces.


\section{Additional Remarks}

A potential drawback in sharing X.509 certificates for both transport security and access control usage is that policy meta-data is then made public. For some cases, policies or name-spaces themselves may be sensitive. For this, the policies would either need to be obfuscated, or transmitted via a secondary certificate after establishing a secure transport tunnel. The first adds the challenge of keeping the un-obscurification method secret, and the later adds complexity in challenging yet another certificate's ownership, signature verification, and CA chain validity. These issues may be addressed with ROS2 and DDS, but for ROS1, SROS intends to provide reasonable security infrastructure while yet minimizing current user disruptions, such as computational overhead and API breakage.

It should be noted that SROS's implementation, specifically its shimming framework between the ROS API and IP stack, has been designed so that both default methods for secure transport and policy evaluation are plug-ins. This should provide advance users the ability to adapt SROS for their own internal certificate formats or policy interpretation. Additionally, security logging is also under development, standardizing the format like that of AppArmor, enabling more powerful auditing and policy generation tools.

Further project documentation can be found at the ROS.org wiki \footnote{\url{http://wiki.ros.org/SROS}}, along with Dockerized \protect\cite{white15roscon} examples to quickly provide readers a starting point in experimenting with SROS.




\nocite{*}
\bibliographystyle{IEEEtran}
\bibliography{references}

\end{document}